\documentclass[10pt,twocolumn,letterpaper]{article}

\usepackage{stfloats}
\usepackage[pagenumbers]{cvpr} 

\def\1{\mathbf{1_{I\times J}}}

\usepackage{color}
\usepackage{graphicx}
\usepackage{amsmath}
\usepackage{amssymb}
\usepackage{booktabs}
\usepackage{multirow}
\usepackage{tikz}
\usepackage{pgfplots}
\pgfplotsset{compat=1.18}
\usepackage[accsupp]{axessibility}

\definecolor{cvprblue}{rgb}{0.21,0.49,0.74}
\usepackage[pagebackref,breaklinks,colorlinks,allcolors=cvprblue]{hyperref}

\usepackage[capitalize]{cleveref}
\crefname{section}{Sec.}{Secs.}
\Crefname{section}{Section}{Sections}
\Crefname{table}{Table}{Tables}
\crefname{table}{Tab.}{Tabs.}

\def\paperID{20475} 
\def\confName{CVPR}
\def\confYear{2026}

\title{Balanced Hierarchical Contrastive Learning with Decoupled Queries for Fine-grained Object Detection in Remote Sensing Images}

\author{Jingzhou Chen$^1$\thanks{Equal contribution.}, Dexin~Chen$^1$\footnotemark[1], Fengchao~Xiong$^1$\thanks{Corresponding author.}, Yuntao Qian$^2$, Liang~Xiao$^1$\footnotemark[2]\\
     $^1$School of Computer Science and Engineering, Nanjing University of Science and Technology, China\\
     $^2$College of Computer Science, Zhejiang University, China\\
     {\tt\small\{jzchen, chendexin, fcxiong, xiaoliang\}@njust.edu.cn, ytqian@zju.edu.cn}
}

\begin{document}
\maketitle
\begin{abstract}
Fine-grained remote sensing datasets often use hierarchical label structures to differentiate objects in a coarse-to-fine manner, with each object annotated across multiple levels. However, embedding this semantic hierarchy into the representation learning space to improve fine-grained detection performance remains challenging. Previous studies have applied supervised contrastive learning at different hierarchical levels to group objects under the same parent class while distinguishing sibling subcategories. Nevertheless, they overlook two critical issues: (1) imbalanced data distribution across the label hierarchy causes high-frequency classes to dominate the learning process, and (2) learning semantic relationships among categories interferes with class-agnostic localization. To address these issues, we propose a balanced hierarchical contrastive loss combined with a decoupled learning strategy within the detection transformer (DETR) framework. The proposed loss introduces learnable class prototypes and equilibrates gradients contributed by different classes at each hierarchical level, ensuring that each hierarchical class contributes equally to the loss computation in every mini-batch. The decoupled strategy separates DETR's object queries into classification and localization sets, enabling task-specific feature extraction and optimization. Experiments on three fine-grained datasets with hierarchical annotations demonstrate that our method outperforms state-of-the-art approaches.
\end{abstract}    
\section{Introduction}\label{sec:intro}

\begin{figure}
    \centering
    \includegraphics[width=1.0\linewidth]{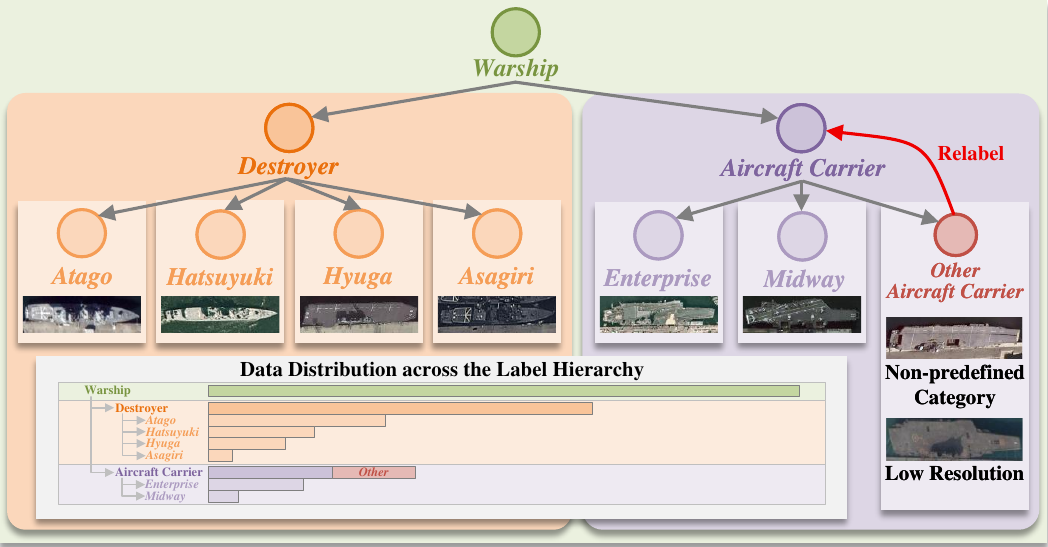}
    \caption{Illustration of the hierarchical label structure in remote sensing datasets. Categories are organized in a coarse-to-fine manner, with leaf nodes representing fine-grained categories. Embedding this structure into the representation learning process poses two primary challenges: (a) data imbalances across and within hierarchical levels, and (b) the conflict between semantically grouping objects under the same parent class while preserving their distinct spatial locations. Additionally, this structure enables flexible reassignment of ambiguous objects to appropriate parent categories.}
    \label{fig:motivation}
\end{figure}

Fine-grained object detection aims to distinguish specific subcategories within broader categories. Existing fine-grained detection datasets~\cite{liu2017high,hou2020fusar,ShipRSImageNet,FAIR1M} typically employ hierarchical label structures to organize numerous fine-grained categories in a coarse-to-fine manner, with sibling classes grouped under a common parent and leaf nodes representing fine-grained categories, as illustrated in Fig.~\ref{fig:motivation}. These hierarchical annotations mirror the recognition process of human experts, facilitating the discrimination of fine-grained categories based on prior hierarchical taxonomy.
Notably, these datasets categorize ambiguous or uncertain objects into special ``other'' classes, such as ``other aircraft carrier''. Previous studies~\cite{ReDet,LSKNet,PETDet,PCLDet,SFRNet,PETDet,SAFPN+APCL,DRNet} treat ``other'' classes as mutually exclusive categories, ignoring their relationships to parent categories. In contrast, hierarchical label structures offer more flexible reassignment of instances from ``other'' classes to corresponding coarse-grained categories, mitigating fine-grained uncertainty while preserving semantic information at the coarse level. The enhanced discriminability and flexibility offered by hierarchical label structures motivate their integration into the representation learning space of detectors.

Recent studies in object detection~\cite{zhang2022hierarchical,zhang2024class,doan2024hyp} and fine-grained classification~\cite{zhang2022use,gupta2023class,10994992} highlight the use of hierarchical label structures for supervised discriminative representation learning. These methods leverage supervised contrastive learning (SCL)~\cite{SCL}, which attracts instances from the same category and repels those from different categories in the embedding space. When adapted to a class hierarchy, this mechanism imposes a dual objective: grouping instances from the same parent class at the coarse-grained level and differentiating sibling subclasses at the fine-grained level.
However, hierarchical label trees are inherently imbalanced. Higher-level nodes accumulate more samples than their descendants, and asymmetric branching leads to skewed distributions among sibling classes, as illustrated in Fig.~\ref{fig:motivation}. Existing studies often overlook data imbalances across and within hierarchical levels, which hinders the model's ability to learn discriminative features for rarer classes.

Another critical challenge is modeling hierarchical semantics for object classification without compromising class-agnostic bounding box regression. Existing detectors usually feed shared representations into task-specific prediction branches for classification and localization. Nevertheless, this unified feature extraction limits the integration of task-specific properties into representation learning without interfering with other tasks. In hierarchical semantic modeling, objects sharing the same parent label can be grouped at the coarse-grained level. Directly applying this grouping to shared representations may cause bounding boxes to cluster together, while they must remain distinct for accurate localization. In other words, objects belonging to the same parent class should exhibit semantic similarity, while preserving distinct spatial locations.

To address the data imbalance, we propose a balanced hierarchical contrastive loss (BHCL) that ensures each class in the label hierarchy contributes equally to the loss computation in every mini-batch. Achieving this objective requires two key components. First, all classes must appear in every mini-batch. However, stochastic mini-batch training does not guarantee the presence of instances from each class, particularly rare classes with few samples. To mitigate this, BHCL introduces a learnable prototype for each class in the label hierarchy as extra instances, enabling all class prototypes to participate in the loss computation for every training batch. Second, each hierarchical class should contribute equally to optimization. The original SCL loss only normalizes the contributions of positive classes by averaging their instances in the numerator, resulting in biased contributions among negative classes in the denominator~\cite{zhu2022balanced}. In contrast, when computed at each hierarchical level, BHCL modifies the SCL loss by further equilibrating the gradients contributed by different negative classes.

To resolve the conflict between grouping objects semantically and separating them spatially, we introduce a decoupled learning strategy within the detection transformer (DETR) framework~\cite{DETR,Deformable-DETR,dai2022ao2,zeng2024ars,OrientedFormer}. In standard DETR, learnable object queries serve as shared representations for generating both classification and localization predictions, which can lead to interference between these tasks when modeling hierarchical semantics.
Our decoupled strategy mitigates this by separating the queries into classification and localization sets, each specialized for its respective task and fed into the corresponding prediction branch. These two types of queries are first aligned with self-attention modules and then refined by dedicated cross-attention modules and feed-forward networks to extract task-specific features from the input images. Finally, the classification queries are projected into the embedding space and supervised by BHCL to effectively model hierarchical semantics.

In conclusion, the main contributions of this paper can be summarized as follows:
\begin{itemize}
    \item To enhance fine-grained object detection, we propose a method that integrates hierarchical label structures into DETR-based detectors. At the core of our approach is a novel balanced hierarchical contrastive loss, which addresses data imbalance across the label hierarchy.
    \item We introduce a simple yet effective decoupled strategy that resolves the optimization conflict between classification and localization in hierarchical semantic modeling, thereby enabling the effective application of our BHCL.
    \item We demonstrate the generality of our proposed contributions by integrating them into two DETR-based architectures, where they deliver consistent performance gains. Our resulting models set new state-of-the-art benchmarks on three fine-grained remote sensing (RS) datasets with hierarchical annotations.
\end{itemize}
\section{Related Work}\label{sec:related}
\subsection{Fine-grained Object Detection in RS}
Fine-grained remote sensing detection datasets often use hierarchical label structures to organize large-scale object categories. For synthetic aperture radar (SAR) datasets, FAIR-CSAR-V1.0~\cite{wu2024fair} classifies ships and aircraft into 5 major classes and 22 subclasses. Similarly, FUSAR-Ship~\cite{hou2020fusar} includes high-resolution SAR images from the Gaofen-3 satellite and categorizes ship targets into 15 parent classes and 98 fine-grained subclasses. For optical datasets, FAIR1M~\cite{FAIR1M} organizes object categories into a two-level hierarchy comprising 5 coarse-grained and 37 fine-grained classes, while ShipRSImageNet~\cite{ShipRSImageNet} structures ship targets into a four-level hierarchy. Additionally, due to factors such as sensor differences, resolution variations, and weather interference, these datasets typically assign hard-to-distinguish objects to ambiguous ``other'' categories.

Recent advancements in fine-grained remote sensing object detection have primarily focused on refining core detection components. Li et al.~\cite{PETDet} enhanced the two-stage detection pipeline with an anchor-free, quality-oriented proposal network for generating high-quality proposals. Xie et al.~\cite{DRNet} proposed generating discriminative representations through a fine-grained feature refinement branch supervised by a confusion-minimized loss. Ouyang et al.~\cite{PCLDet} introduced a prototypical contrastive learning strategy that leverages class prototypes. Li et al.~\cite{SAFPN+APCL} designed a spatial-aligned feature pyramid network to address spatial misalignment in multiscale feature fusion. Although these approaches achieved competitive performance, they overlooked the hierarchical label structures inherent in fine-grained datasets.

\subsection{Hierarchical Object Detection}
Object detection based on hierarchical label structures requires modeling the semantic relationships among hierarchical categories, in addition to localizing object instances. Remote sensing images typically cover large areas, with targets occupying a relatively small portion of the overall area. Leveraging prior knowledge from the hierarchical label structure, Wu et al.~\cite{wu2021hierarchical} first detected potential regions where targets may be located, followed by further object detection within these localized areas. For example, aircraft targets usually appear in airport areas, while ship targets are commonly found near ports. Shin et al.~\cite{shin2020hierarchical} first employed a generic detector to extract object regions, which were then fed into a subsequent hierarchical classification network to predict the hierarchical categories of the extracted objects.

Addressing the hierarchical few-shot object detection problem, Zhang et al.~\cite{zhang2022hierarchical} utilized beam search to obtain prediction probabilities for categories along multiple paths in the label hierarchy, thereby mitigating the error propagation from parent to child nodes during prediction. To tackle few-shot ship detection, Zhang et al.~\cite{zhang2024class} adopted a two-stage strategy: first, a detector extracted all ``ship'' targets; subsequently, learnable prototypes were established for each category in the label hierarchy, with target categories determined by nearest-neighbor classification. To perform open-world object detection, Doan et al.~\cite{doan2024hyp} used a two-level hierarchy to determine whether an unknown target belonged to an existing parent class by calculating the distance between unknown target representations and parent class vectors in hyperbolic space.
In contrast to the aforementioned methods, our proposed approach improves fine-grained object detection performance by integrating hierarchical label structures into the end-to-end DETR architecture, without splitting the localization and classification tasks into two separate stages.
\section{Proposed Method}\label{sec:methods}

\subsection{Overview}
\begin{figure*}
    \centering
    \includegraphics[width=0.94\linewidth]{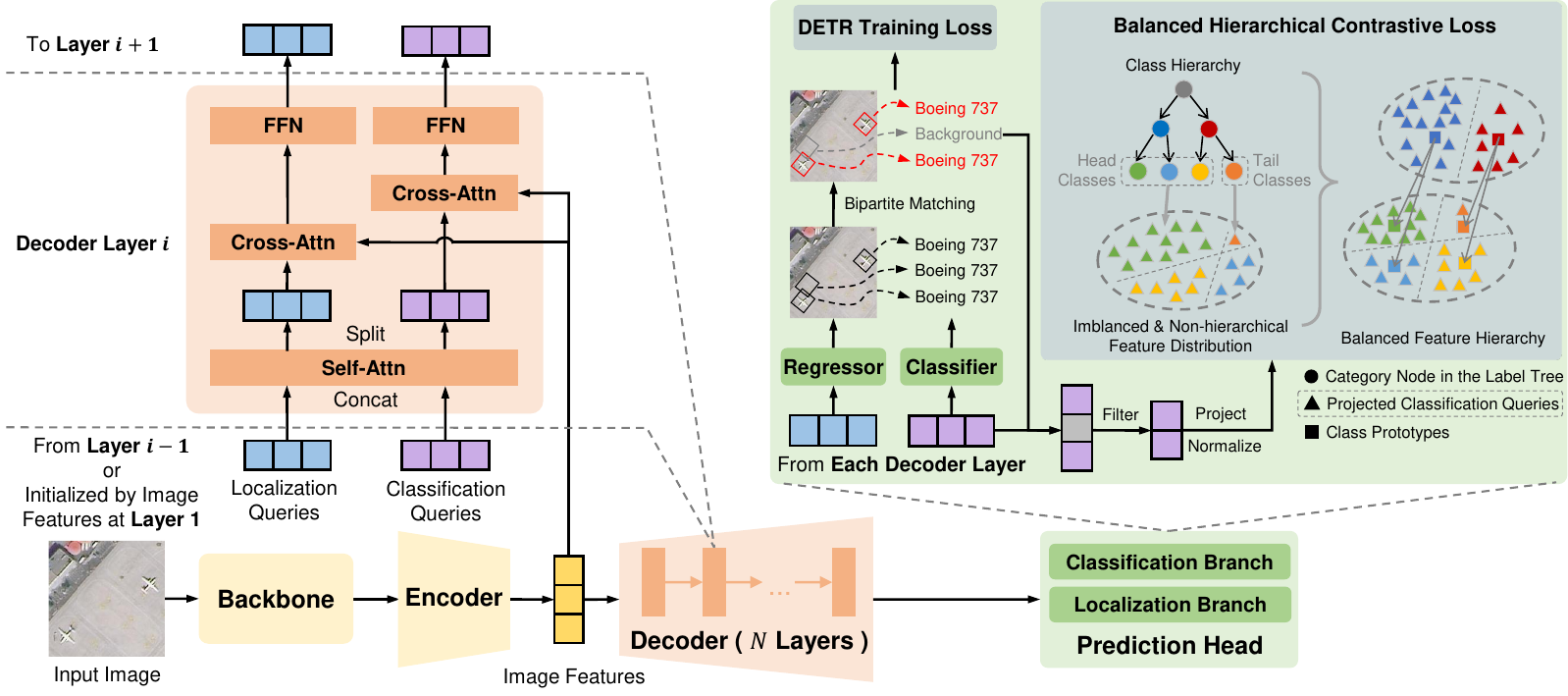}
    \caption{Architecture of the proposed method. A backbone network and an encoder are employed to extract and refine image features, respectively. These refined features initialize a set of learnable object queries, which are then decoupled into classification and localization queries. At each decoder layer, these queries first interact through a shared self-attention module, followed by two parallel streams for classification and localization tasks. Each task-specific stream contains a cross-attention module and a feed-forward network. Finally, dedicated branches in the prediction head generate class predictions from the refined classification queries and regress bounding boxes from the localization queries. During training, bipartite matching assigns ground truth to the predictions, and the $L_1$ loss, IoU loss, classification loss, and the proposed balanced hierarchical contrastive loss (BHCL) are computed for model optimization. BHCL embeds the class hierarchy into the object representation space within the DETR framework, introducing learnable prototypes for each class in the hierarchy to ensure balanced contributions from all classes to the loss computation in every mini-batch.}
    \label{fig:architecture}
    \vspace{-0.5em}
\end{figure*}

The overall architecture consists of four components: the backbone, transformer encoder, transformer decoder, and prediction head, as illustrated in Fig.~\ref{fig:architecture}. The backbone network processes an input image $x \in\mathbb{R}^{H \times W \times B}$, where $H$, $W$, and $B$ represent the height, width, and number of spectral bands, respectively. It extracts feature maps from the image, which are then flattened into a sequence of image tokens, with 2D positional encoding added. The encoder utilizes self-attention modules to capture long-range dependencies between the image tokens, processing them layer by layer. The output from the final encoder layer is denoted as $z\in\mathbb{R}^{M \times d}$, where $M$ denotes the number of tokens and $d$ represents their dimensionality.

The transformer decoder processes two types of inputs: $z$ from the encoder and a set of learnable object queries. These queries act as proposals to predict candidate objects in the input image.
Each decoder layer comprises a self-attention module, a cross-attention module, and a feed-forward network. At each decoder layer, the object queries interact with $z$ through the cross-attention module to integrate object features. After the decoder layers, the updated queries are fed into the prediction head to generate object predictions, including the category probability distribution and the rotated bounding box coordinates for each object. During training, bipartite matching~\cite{DETR} is used to assign each ground truth object to a unique query by minimizing the following cost function:
\begin{equation}
\mathcal{C}_{\text{match}}=\lambda'_{\text{cls}}\mathcal{C}_{\text{cls}}+\lambda'_{\text{iou}}\mathcal{C}_{\text{iou}}+\lambda'_{\text{L}_1}\mathcal{C}_{\text{L}_1},
\end{equation}
where $\mathcal{C}_{\text{cls}}$ is the Focal loss~\cite{lin2017focal} for classification, $\mathcal{C}_{\text{iou}}$ is the Rotate IoU loss~\cite{zhou2019iou}, and $\mathcal{C}_{\text{L}_1}$ is the $L_1$ loss for regression. $\lambda'_{\text{cls}},\;\lambda'_{\text{iou}},\;\lambda'_{\text{L}_1}$ are the coefficients for their respective costs. Queries that do not match any objects are assigned to the background.

After bipartite matching, the prediction results of the object queries and their matched ground truth are used to form the training loss for optimizing the entire model:
\begin{equation}\label{loss:total}
\mathcal{L}_{\text{total}}=\lambda_{\text{BHCL}}\mathcal{L}_{\text{BHCL}}+\lambda_{\text{cls}}\mathcal{L}_{\text{cls}}+\lambda_{\text{iou}}\mathcal{L}_{\text{iou}}+\lambda_{\text{L}_1}\mathcal{L}_1,
\end{equation}
where $\mathcal{L}_{\text{cls}}$ is the Focal loss, $\mathcal{L}_{\text{iou}}$ is the Rotate IoU loss, and $\mathcal{L}_{\text{L}_1}$ is the $L_1$ loss. $\lambda_{\text{cls}},\;\lambda_{\text{iou}},\;\lambda_{\text{L}_1}$ are the weights assigned to each loss term. Additionally, $\mathcal{L}_{\text{BHCL}}$ represents our proposed balanced hierarchical contrastive loss, which explicitly models hierarchical semantics. The detailed formulation of $\mathcal{L}_{\text{BHCL}}$ is elaborated in the subsequent section.

\subsection{Decoupled Learning of Object Queries}
DETR establishes direct correspondences between object representations and semantic categories by assigning learnable object queries to each ground-truth category via bipartite matching. These queries act as candidate proposals and interact with image features through cross-attention modules to generate classification and localization predictions. Modeling hierarchical semantics requires grouping objects under a common parent category. Imposing this grouping on object queries may lead to interference between classification and localization tasks. Therefore, hierarchical semantics should be integrated into object query classification without affecting class-agnostic bounding box learning. This ensures that spatial localization remains independent of hierarchical relationships between categories.

In our approach, the object queries are decoupled into classification queries $Q_{\text{cls}}\in\mathbb{R}^{N\times d}$ and localization queries $Q_{\text{loc}}\in\mathbb{R}^{N\times d}$, where $N$ is the number of object queries. At each decoder layer, $Q_{\text{cls}}$ and $Q_{\text{loc}}$ are first concatenated along the hidden dimension $d$ and passed into the shared self-attention (Self-Attn) module to align classification and localization information for the same objects:
\begin{equation}
\begin{aligned}
Q_{\text{cls}},Q_{\text{loc}}=\text{Split}(\text{Self-Attn}(\text{Concat}(Q_{\text{cls}},Q_{\text{loc}})).
\end{aligned}
\end{equation}
Both $Q_{\text{cls}}$ and $Q_{\text{loc}}$ interact with $z$ through their respective cross-attention (Cross-Attn) modules and feed-forward networks (FFN) to extract task-specific image features:
\begin{equation}
\begin{aligned}
Q'_{\text{cls}}&=\text{FFN}_1(\text{Cross-Attn}_1(Q_{\text{cls}}, z)), \\
Q'_{\text{loc}}&=\text{FFN}_2(\text{Cross-Attn}_2(Q_{\text{loc}}, z)).
\end{aligned}
\end{equation}
Finally, the prediction head, consisting of separate classification and localization branches, processes the specialized queries from the last decoder layer to generate class and bounding box predictions, respectively. The proposed BHCL is applied exclusively to $Q'_{\text{cls}}$.

\subsection{Hierarchical Semantic Modeling}
Remote sensing datasets typically utilize hierarchical label trees to organize numerous fine-grained object categories. Each object is associated with multiple category labels along the path from the root to the leaf node in the hierarchical label tree. As a result, objects can be grouped or differentiated at different hierarchical levels based on their category labels at each corresponding level. We propose integrating the hierarchical label tree into the representation learning of object queries within transformer-based detectors to improve the model's discriminative ability among fine-grained leaf nodes.
Supervised contrastive learning provides a foundation by leveraging label information to pull together object queries of the same category while pushing apart those from different categories. Our hierarchical semantic modeling extends this framework by incorporating the hierarchical label tree and introducing decoupled classification queries.

\subsubsection{Hierarchical Contrastive Learning}
Given the decoupled object queries, each classification query is assigned a ground truth category label after bipartite matching. The projection head then maps the classification queries into a lower-dimensional representation space and applies normalization to the projected representations. The supervised contrastive loss (SCL) for a mini-batch is defined as follows:
\begin{equation}
\begin{aligned}
\mathcal{L}_{\text{SCL}}=-\frac{1}{|I|}\sum_{i\in I}\frac{1}{|P(i)|}\sum_{p\in P(i)}\mathcal{L}_{\text{pair}}(i,p),
\end{aligned}
\end{equation}
where $P(i)$ represents the index set of classification queries that share the same category as the $i$-th query, excluding the $i$-th query itself, and $I$ denotes the index set of all classification queries that are matched to foreground objects (i.e., non-background) in the mini-batch. The pair contrastive loss $\mathcal{L}_{\text{pair}}(i,p)$ for a positive pair is computed as follows:
\begin{equation}
\begin{aligned}
\mathcal{L}_{\text{pair}}(i,p)=-\log\frac{\exp(f_i\cdot f_p/\tau)}{\sum_{a\in I\backslash\{i\}}\exp(f_i\cdot f_a/\tau)},
\end{aligned}
\end{equation}
where $f_i$ and $f_p$ are the $\mathbb{\ell}_2$-normalized projected classification queries belonging to the same category, $\tau$ denotes the temperature coefficient, and $\cdot$ represents the inner product between two vectors.

Hierarchical contrastive learning organizes positive and negative pairs of normalized classification queries based on their category labels at each level of the hierarchical label tree and calculates the corresponding supervised contrastive loss. The hierarchical contrastive loss (HCL) is then formulated as a weighted sum of the contrastive losses across all levels:
\begin{equation}
\begin{aligned}
\mathcal{L}_{\text{HCL}}&=-\frac{1}{|I|}\sum_{i\in I}\sum_{l=1}^{L}\frac{\lambda_l}{|P_l(i)|}\sum_{p\in P_l(i)}\mathcal{L}_{\text{pair}}(i,p),
\end{aligned}
\end{equation}
where $L$ represents the number of levels in the hierarchical label tree, $P_l(i)$ denotes the index set of classification queries sharing the same ancestor category at level $l$ as the $i$-th query, and $\lambda_l=\exp(\frac{1}{L+1-l})/\sum_{l'=1}^{L}\exp(\frac{1}{L+1-l'})$ is the penalty term~\cite{zhang2022use} for level $l$. Levels closer to the leaf nodes are assigned higher penalty weights, encouraging the model to focus more on distinguishing fine-grained categories.
It is worth noting that the top level of the hierarchical label tree, which contains only a single root node, is excluded from our computation.

\subsubsection{Prototype-Based Class Balancing}
In a hierarchical label tree, the distribution of training samples is often imbalanced among fine-grained leaf nodes. Additionally, the tree structure itself is typically imbalanced, with intermediate nodes having a varying number of child nodes. Consequently, these two factors lead to imbalanced sample distributions across all category nodes within the label hierarchy. In hierarchical contrastive learning, high-frequency classes tend to dominate the representation learning process when dealing with such imbalanced data, resulting in suboptimal performance across all classes. To address this issue, the key idea is to ensure that each class contributes equally to the optimization of the hierarchical contrastive loss in every mini-batch. However, not all categories appear in a single mini-batch, and the number of samples from each category within the mini-batch may vary.

To overcome these challenges, we first establish a learnable class prototype bank $\mathcal{M}\in\mathbb{R}^{C\times d'}$, where $C$ denotes the number of category nodes in the label tree excluding the root node, and $d'$ is the same dimension as the projected classification queries. Then, we average the instances of each class in the mini-batch to ensure that each class has an equal contribution to the optimization process. Accordingly, the previous pair contrastive loss $\mathcal{L}_{\text{pair}}$ is modified as follows:
\begin{equation}
\begin{aligned}
\mathcal{L}^{b}_\text{pair}(l,i,p)=-\log\frac{\exp(f_i\cdot f_p/\tau)}{\underset{c\in C_l}{\sum}\frac{1}{|I'_c|}\underset{a\in I'_c\backslash\{i\}}{\sum}\exp(f_i\cdot f_a/\tau)},
\end{aligned}
\end{equation}
where $C_l$ denotes the set of all categories located at level $l$, and $I'_c=I_c\cup\{\mathcal{M}(c)\}$. $\mathcal{M}(c)$ is the class prototype corresponding to category $c$. $I_c$ represents the index set of projected classification queries belonging to category $c$. Note that $I_c$ may be an empty set when no instances of category $c$ are present in the mini-batch. The balanced hierarchical contrastive loss is defined as:
\begin{equation}
\begin{aligned}
\mathcal{L}_{\text{BHCL}}=-\frac{1}{|I|}\sum_{i\in I}\sum_{l=1}^{L}\frac{\lambda_l}{|P'_l(i)|}\sum_{p\in P'_l(i)}\mathcal{L}^{b}_\text{pair}(l,i,p),
\end{aligned}
\end{equation}
where $P'_l(i)=P_l(i)\cup\{\mathcal{M}(l,i)\}$, and $\mathcal{M}(l,i)$ refers to the class prototype corresponding to the ancestor category of the $i$-th query at level $l$.

The balanced hierarchical contrastive loss, $\mathcal{L}_{\text{BHCL}}$, is used in the training loss defined in Eq.~\ref{loss:total} to jointly optimize the parameters of the detector. To enhance hierarchical semantic modeling among classification queries, $\mathcal{L}_{\text{BHCL}}$ is applied at each decoder layer. Each class prototype $\mathcal{M}_c\in\mathbb{R}^{d'}$ is updated using exponential moving average (EMA) as follows:
\begin{equation}
\begin{aligned}
\mathcal{M}_{c}\leftarrow(1-\epsilon^{L-l})\mathcal{M}_c+\epsilon^{L-l}\bar{f}_c,
\end{aligned}
\end{equation}
where $\epsilon$ is the momentum coefficient, and $\bar{f}_c$ is the average of the projected classification queries matched with category $c$. It is worth mentioning that, for coarse-grained category nodes at intermediate levels, their prototypes are updated by averaging the projected classification queries matched with themselves and all their descendant classes.

Regarding datasets with hierarchical category annotations, ambiguous objects are categorized into a special ``Other'' category, such as labeling an uncertain aircraft carrier as ``Other Aircraft Carrier.'' In contrast to existing methods that treat the ``Other'' category as a mutually exclusive fine-grained class, the proposed method assigns these ambiguous objects to more general parent classes and applies hierarchical contrastive learning. In other words, the hierarchical contrastive loss utilizes training instances that are distributed not only at fine-grained leaf nodes but also at intermediate nodes of the hierarchical label tree.
\section{Experiments}\label{sec:exper}

\subsection{Datasets}
Experiments are conducted on three fine-grained datasets with class hierarchies, including ShipRSImageNet~\cite{ShipRSImageNet}, FAIR1M-v1.0, and FAIR1M-v2.0~\cite{FAIR1M}.
ShipRSImageNet contains 3,435 images with spatial resolution ranging from 0.12m to 6m and image size ranging from 930$\times$930 pixels to 1024$\times$1024 pixels, comprising a total of 17,573 instances. It consists of 41 fine-grained ship categories organized into a four-level class hierarchy by adding 8 coarse-grained categories as ancestors. In addition, there are 8 \textit{Other*} categories, such as \textit{Other Ship}, which cover instances that cannot be further identified due to low image resolution or those that do not belong to any predefined fine-grained categories.

FAIR1M-v1.0 and FAIR1M-v2.0 share a training set of 16,488 images. However, FAIR1M-v2.0 expands the test set of FAIR1M-v1.0 from 8,137 images to 18,021 images and introduces an additional validation set of 8,287 images. The spatial resolution of these images ranges from 0.3m to 0.8m, with sizes varying between 1000$\times$1000 pixels and 10000$\times$10000 pixels. Both FAIR1M-v1.0 and FAIR1M-v2.0 contain 34 fine-grained categories organized into a two-level class hierarchy by adding 5 coarse-grained ancestor categories. Similar to ShipRSImageNet, there are 3 \textit{Other*} categories in both FAIR1M-v1.0 and FAIR1M-v2.0.

\subsection{Implementation Details}
The proposed method is applicable to any DETR-based detector, and we adopt two recently proposed DETR-based detectors, OrientedFormer \cite{OrientedFormer} and RHINO \cite{RHINO}, as baselines. The weight of the balanced hierarchical contrastive loss, $\lambda_{\text{BHCL}}$, is set to 0.6 based on our ablation study. The temperature coefficient $\tau$ and momentum coefficient control factor $\epsilon$ are empirically set to 0.1 without further tuning. The models are optimized using AdamW \cite{AdamW} with a learning rate of $5\times10^{-5}$ and a batch size of 8. Random flip and random shift are applied to generate two augmented views of each input image. All experiments are conducted on 4 NVIDIA RTX 4090 GPUs. Unlike previous studies, which treat \textit{Other*} categories as mutually exclusive fine-grained categories, we reassign instances from \textit{Other*} categories to their corresponding parent categories. For example, instances from \textit{Other Aircraft Carrier} are reassigned to \textit{Aircraft Carrier}.

In our experiments, we report rotated COCO-style metrics~\cite{MMRotate}, including $AP_{50}$, $AP_{75}$, and $AP_{50:95}$. Since the test set annotations for ShipRSImageNet are not publicly available and no evaluation server is provided, models are trained on the training set and evaluated on the validation set. In contrast, for FAIR1M-v1.0 and v2.0, although test annotations are also unavailable, an official evaluation server enables submission of inference outputs for verified $AP_{50}$ results, which are subsequently reported.

\begin{table}\small
\caption{Ablation Study on the Effectiveness of Key Components: Decoupling Strategy, Hierarchical Contrastive Loss (HCL), and Balanced Hierarchical Contrastive Loss (BHCL). OF denotes OrientedFormer for brevity.\label{tab:EffectivenessOfEachComponents}}
\centering
\begin{tabular}{lccc}
\toprule
& $AP_{50}$ & $AP_{75}$ & $AP_{50:95}$ \\
\midrule
\textit{ShipRSImageNet} \\
\midrule
RHINO & 78.90 & 72.70 & 59.78 \\
RHINO + Decoupling & 78.70 & 73.00 & 60.99 \\
RHINO + Decoupling + HCL & \textbf{79.90} & 73.70 & 61.24 \\
RHINO + Decoupling + BHCL & 79.80 & \textbf{74.30} & \textbf{61.41} \\
\midrule
OF & 78.60 & 73.80 & 63.17 \\
OF + Decoupling & 79.60 & 73.80 & 63.60 \\
OF + Decoupling + HCL & \textbf{80.50} & 74.00 & 64.12 \\
OF + Decoupling + BHCL & 80.40 & \textbf{74.90} & \textbf{64.32} \\
\midrule
\textit{FAIR1M-v1.0} \\
\midrule
OF & 41.31 & - & - \\
OF + Decoupling & 41.38 & - & - \\
OF + Decoupling + HCL & 41.14 & - & - \\
OF + Decoupling + BHCL & \textbf{41.66} & - & - \\
\bottomrule
\end{tabular}
\end{table}

\subsection{Ablation Studies}
\subsubsection{Effectiveness of Each Component across Baselines}
The proposed method consists of three key components: the decoupled learning strategy (decoupling), the hierarchical contrastive loss (HCL) that applies supervised contrastive loss across hierarchical levels, and the balanced hierarchical contrastive loss (BHCL) that modifies HCL with the prototype-based class balancing mechanism. Table \ref{tab:EffectivenessOfEachComponents} reports ablation results for each component on two baseline detectors, RHINO and OrientedFormer, across two datasets. As shown in Table \ref{tab:EffectivenessOfEachComponents}, the decoupled strategy yields improvements of $+1.21$ $AP_{50:95}$ on RHINO and $+0.43$ $AP_{50:95}$ on OrientedFormer, demonstrating that decoupling enables the classification and localization tasks to focus on task-specific feature learning. Furthermore, applying HCL to the classification queries from each decoder layer achieves gains of $+0.25$ $AP_{50:95}$ on RHINO and $+0.52$ $AP_{50:95}$ on OrientedFormer, confirming its effectiveness in enhancing fine-grained category discrimination. Finally, incorporating the balancing mechanism into HCL delivers additional gains of $+0.17$ $AP_{50:95}$ on RHINO and $+0.20$ $AP_{50:95}$ on OrientedFormer. These ablation results collectively validate the contribution of each component. Given the consistently superior performance of OrientedFormer, it is adopted as the baseline for all subsequent experiments.

\subsubsection{Effectiveness of Each Component across Datasets}
\label{sec:exper_differentDatasets}
To further evaluate the generalizability of the proposed method, we conduct additional experiments on the FAIR1M-v1.0 dataset. As shown in Table \ref{tab:EffectivenessOfEachComponents}, the hierarchical contrastive loss results in performance degradation on the FAIR1M-v1.0 dataset (even worse than the baseline), while the balanced hierarchical contrastive loss consistently improves performance ($+0.28$ $AP_{50}$). We hypothesize that the FAIR1M-v1.0 dataset exhibits a more pronounced long-tail distribution compared to ShipRSImageNet, causing the hierarchical contrastive loss to be dominated by instances from the head class. This underscores the critical importance of the balancing mechanism in the hierarchical contrastive loss. The verification results of the instance distribution per category in the ShipRSImageNet and FAIR1M training sets are provided in the supplementary material.

\subsubsection{t-SNE Visualization}
\begin{figure}
    \centering
    \includegraphics[width=1.0\linewidth]{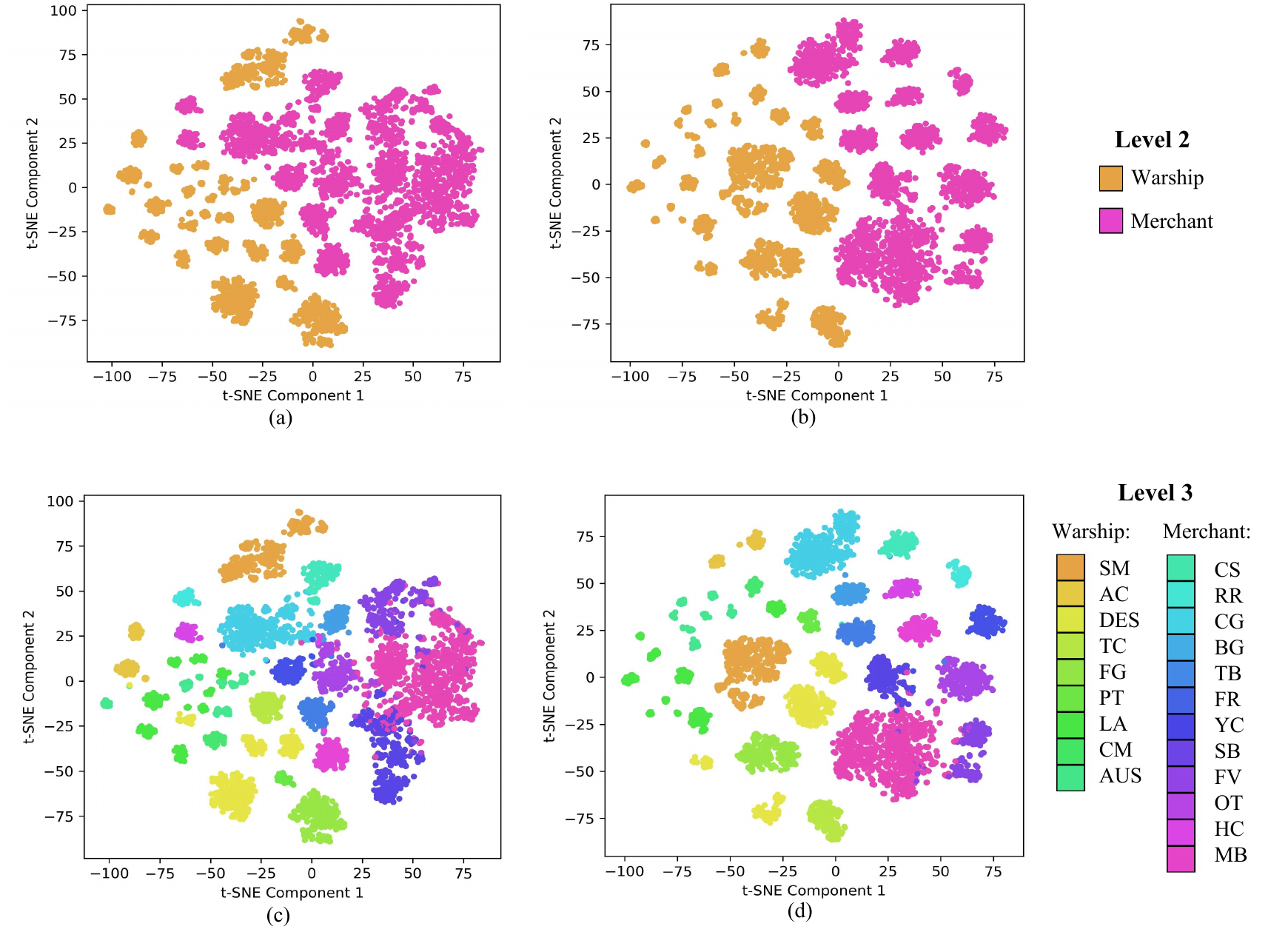}
    \caption{t-SNE Visualization of Classification Queries: (a) OrientedFormer at Level 2, (b) Proposed Method at Level 2, (c) OrientedFormer at Level 3, (d) Proposed Method at Level 3. The full names of the abbreviated subcategories under the coarse-grained categories ``Warship'' and ``Merchant'' are provided in the supplementary material.}
    \label{fig:t-SNE_Result}
\end{figure}

To understand why our proposed loss is effective, we employ t-SNE to project the classification queries into a 2D space, visualizing the distribution of different categories in the feature space across different levels. With the help of BHCL, the model learns hierarchical class representations that align with the class hierarchy. As shown in Fig. \ref{fig:t-SNE_Result}, at level 2, BHCL sharpens the boundary between \textit{Warship} and \textit{Merchant}. Furthermore, at level 3, subcategories under \textit{Warship} and \textit{Merchant} exhibit enhanced separation, indicating improved fine-grained discrimination.

\begin{figure}
    \hspace{0.3cm}
    \begin{tikzpicture}
        \begin{axis}[
            xlabel={$\lambda_{\text{BHCL}}$},
            ylabel={$AP_{50:95}$},
            grid=both,
            width=7.3cm,
            height=3.3cm,
            ymin=62,
            ymax=65,
            xmin=0.4,
            xmax=0.8,
            enlargelimits,
            xlabel style={font=\small},
            ylabel style={font=\small}
        ]
        \addplot[mark=square*, red] coordinates {(0.4, 63.38) (0.5, 63.59) (0.6, 64.32) (0.7, 63.61) (0.8, 62.97)};
        \end{axis}
    \end{tikzpicture}
    \caption{Comparison of Different Weights for Balanced Hierarchical Contrastive Loss.}
    \label{fig:lambdaBHCL}
\end{figure}
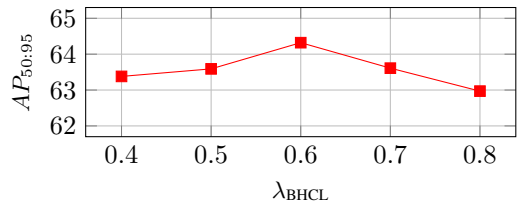

\begin{table}
\caption{Comparison of Different Class Prototype Implementations. None: No additional prototypes; EMA: EMA-updated class prototypes; Cls-Weight: Classifier weights as class prototypes.}\label{tab:PrototypeSettings}
\centering
\begin{tabular}{cccc}
\toprule
Setting & $AP_{50}$ & $AP_{75}$ & $AP_{50:95}$ \\
\midrule
None & 79.30 & 73.80 & 63.34 \\
EMA & \bf{80.40} & 74.90 & \bf{64.32} \\
Cls-Weight & 80.10 & \bf{75.30} & 64.26 \\
\bottomrule
\end{tabular}
\end{table}

\subsubsection{Weight Selection for BHCL}\label{Weight_BHCL}
As shown in Fig. \ref{fig:lambdaBHCL}, the weight of the balanced hierarchical contrastive loss, $\lambda_{\text{BHCL}}$, is evaluated at intervals of $0.1$, identifying $\lambda_{\text{BHCL}}=0.6$ as the optimal value. Accordingly, this value is employed in subsequent experiments.

\subsubsection{Class Prototype Implementations}
The balanced hierarchical contrastive loss adopts EMA-updated class prototypes as complementary instances. To investigate the impact of prototype selection, we evaluate two alternative implementations. The first computes the loss without class prototypes (denoted as \textit{None}). The second uses the classifier weight $W\in\mathbb{R}^{d\times C}$ from the prediction head as class prototypes (denoted as \textit{Cls-Weight}), where each column corresponds to a specific category in the class hierarchy. As shown in Table \ref{tab:PrototypeSettings}, the \textit{None} variant incurs significant performance degradation ($-0.92$ $AP_{50:95}$ compared to \textit{Cls-Weight} and $-0.98$ $AP_{50:95}$ compared to \textit{EMA}). This is likely because class prototypes ensure that the representations of tail classes appear frequently in each batch, maintaining their consistent contribution to the computation of BHCL. Additionally, the \textit{EMA} implementation marginally outperforms \textit{Cls-Weight} ($+0.06$ $AP_{50:95}$).

\setlength{\tabcolsep}{5pt}
\begin{table*}\footnotesize
\caption{Comparison with State-of-the-Art Methods on ShipRSImageNet and FAIR1M-v1.0/v2.0. Results for all compared methods, except RHINO and OrientedFormer, are taken from \cite{SAFPN+APCL} for ShipRSImageNet and from \cite{DRNet} for FAIR1M-v1.0/v2.0. All methods use a ResNet-50 backbone, and input images are resized to $1024\times1024$ pixels.}\label{tab:SOTA}
\centering
\begin{tabular}{ccccccccccc}
\toprule
\multirow{2}*{\vspace{-4pt}\bf{ShipRSImageNet}} & Method & ReDet & ORCNN & LSKNet & SAFPN+APCL & PCLDet & PETDet & OrientedFormer & RHINO & \bf{Ours} \\
\cmidrule{2-11}
~ & $AP_{50:95}$ & 45.1 & 59.5 & 61.0 & 62.7 & 61.6 & 58.7 & 63.2 & 59.8 & \bf{64.3} \\
\midrule
\multirow{2}*{\vspace{-4pt}\bf{FAIR1M-v1.0}} & Method & FRCNN & ORCNN & RoITrans & SFRNet & PCLDet & PETDet & OrientedFormer & DRNet & \bf{Ours} \\
\cmidrule{2-11}
~ & $AP_{50}$ & 36.83 & 38.85 & 38.49 & 40.74 & 40.39 & 40.55 & 41.31 & 40.87 & \bf{41.66} \\
\midrule
\multirow{2}*{\vspace{-4pt}\bf{FAIR1M-v2.0}} & Method & FRCNN & ORCNN & RoITrans & SFRNet & PCLDet & PETDet & GVertex & DRNet & \bf{Ours} \\
\cmidrule{2-11}
~ & $AP_{50}$ & 39.37 & 44.29 & 44.11 & 45.68 & 44.66 & 46.10 & 40.33 & 47.04 & \bf{47.53} \\
\bottomrule
\end{tabular}
\end{table*}

\begin{figure}
    \centering
    \includegraphics[width=1.0\linewidth]{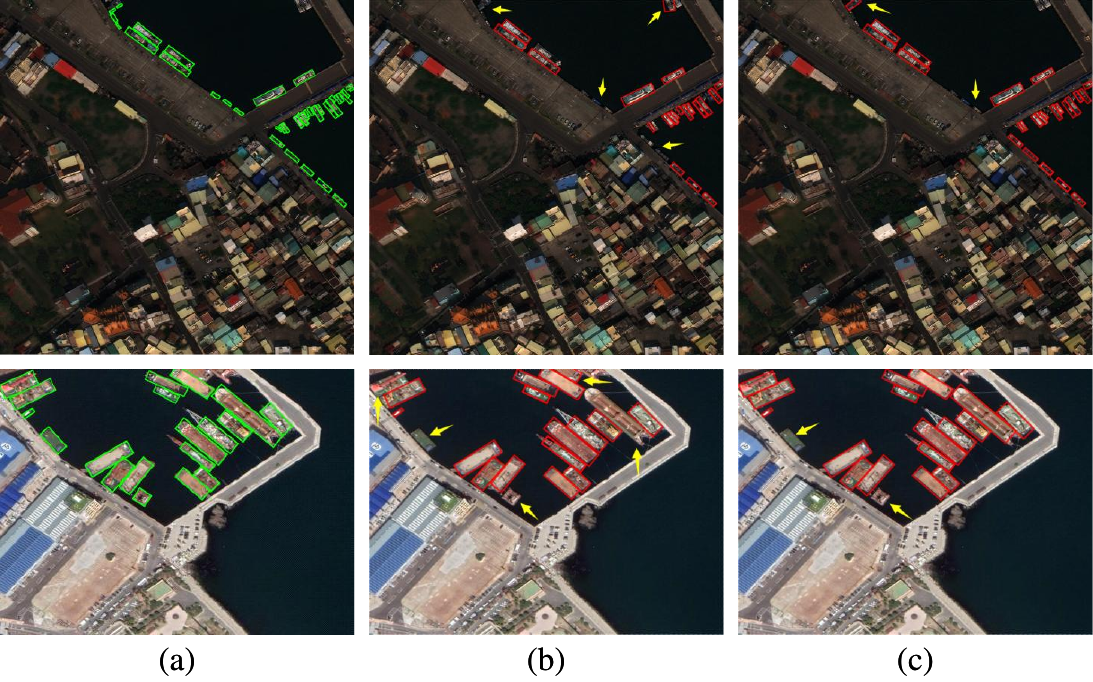}
    \caption{Comparison of Detection Results on \mbox{ShipRSImageNet} (top) and \mbox{FAIR1M-v2.0} (bottom) Validation Sets: (a) Ground Truth, (b) \mbox{OrientedFormer}, (c) Proposed Method (Best Results).}
    \label{fig:detectionResults}
\end{figure}

\subsection{Comparisons With State-of-the-art Methods}
\label{sec:sota_comparison}
This section compares the proposed method with state-of-the-art approaches, including both CNN-based and DETR-based detectors, on the ShipRSImageNet, \mbox{FAIR1M-v1.0}, and \mbox{FAIR1M-v2.0} datasets. For CNN-based detectors, we compare against ReDet \cite{ReDet}, ORCNN \cite{ORCNN}, LSKNet \cite{LSKNet}, SAFPN+APCL \cite{SAFPN+APCL}, PCLDet \cite{PCLDet}, PETDet \cite{PETDet}, FRCNN \cite{FRCNN}, RoITrans \cite{RoITrans}, SFRNet \cite{SFRNet}, GVertex \cite{GVertex}, and DRNet \cite{DRNet}. Their results are taken from \cite{SAFPN+APCL} for ShipRSImageNet and from \cite{DRNet} for \mbox{FAIR1M-v1.0/v2.0}. For DETR-based detectors, we evaluate against OrientedFormer \cite{OrientedFormer} and RHINO \cite{RHINO}, with results reproduced using their official open-source implementations. The comparative results are presented in Table~\ref{tab:SOTA}. Additional detailed comparisons are provided in the supplementary material.

On the ShipRSImageNet dataset, our method achieves an $AP_{50:95}$ of $64.3$, outperforming the second-best method, OrientedFormer, by $+1.1$ and establishing a new benchmark. It is worth mentioning that both PCLDet and SAFPN+APCL adopt contrastive learning with class prototypes to enhance inter-class separability, but they compute contrastive loss solely at the fine-grained level without leveraging the class hierarchy. In contrast, our approach clusters class representations across multiple levels to improve their separability, resulting in substantial gains of $+2.7$ $AP_{50:95}$ over PCLDet and $+1.6$ $AP_{50:95}$ over SAFPN+APCL. This underscores that embedding the hierarchical label structure into the representation learning space of detectors improves fine-grained object detection performance.

On the \mbox{FAIR1M-v1.0} dataset, our method achieves an $AP_{50}$ of $41.66$, surpassing the second-best method, OrientedFormer, by $+0.35$ $AP_{50}$. It also outperforms PCLDet by $+1.27$ $AP_{50}$. The recently proposed DRNet designs a fine-grained branch to encode object features into discriminative representations and further enhances discriminative ability through a confusion-minimized loss. Our method exceeds DRNet by $+0.79$ $AP_{50}$, highlighting its advantages in fine-grained object detection.
On the \mbox{FAIR1M-v2.0} dataset, all methods achieve significant performance gains due to the additional validation set for training. Nevertheless, our method achieves the highest $AP_{50}$ of $47.53$ among all compared approaches, surpassing the second-best DRNet by $+0.49$ $AP_{50}$.

\subsection{Visualization of Detection Results}
\label{sec:vis_det_results}
Fig.~\ref{fig:detectionResults} illustrates the detection results of OrientedFormer and the proposed method on the ShipRSImageNet and \mbox{FAIR1M-v2.0} validation sets. Our method misses fewer ship targets, leading to better overall performance. However, as highlighted in the figure, our method still faces limitations in detecting some tiny objects.
\section{Conclusion}\label{sec:con}
In this paper, we propose a novel approach to improve fine-grained object detection by effectively integrating hierarchical label structures into the end-to-end DETR architecture. When modeling hierarchical semantics, our method addresses the challenges of data imbalance and interference between classification and localization tasks by introducing a balanced hierarchical contrastive loss and a decoupled learning strategy. The balanced hierarchical contrastive loss ensures that each class in the label hierarchy contributes equally to the loss computation, while the decoupled strategy separates classification and localization queries, enabling independent optimization for task-specific feature extraction. Experimental results on three fine-grained remote sensing datasets demonstrate the effectiveness of the proposed method, setting new state-of-the-art benchmarks across all datasets. Future work will explore extending our approach to other detection datasets beyond remote sensing and to other vision tasks involving hierarchical categorization, such as instance segmentation and multi-label classification.
{
    \small
    \bibliographystyle{ieeenat_fullname}
    \bibliography{main}
}
\clearpage
\setcounter{page}{1}
\maketitlesupplementary

\section{Hierarchical Annotations of Datasets}
In our experiments, we utilize three fine-grained detection datasets with hierarchical annotations: ShipRSImageNet \cite{ShipRSImageNet}, FAIR1M-v1.0, and FAIR1M-v2.0\cite{FAIR1M}. Details of the fine-grained categories and their corresponding class hierarchies for each dataset are provided below.

\begin{figure*}[!t]
    \centering
    \includegraphics[width=1.0\linewidth]{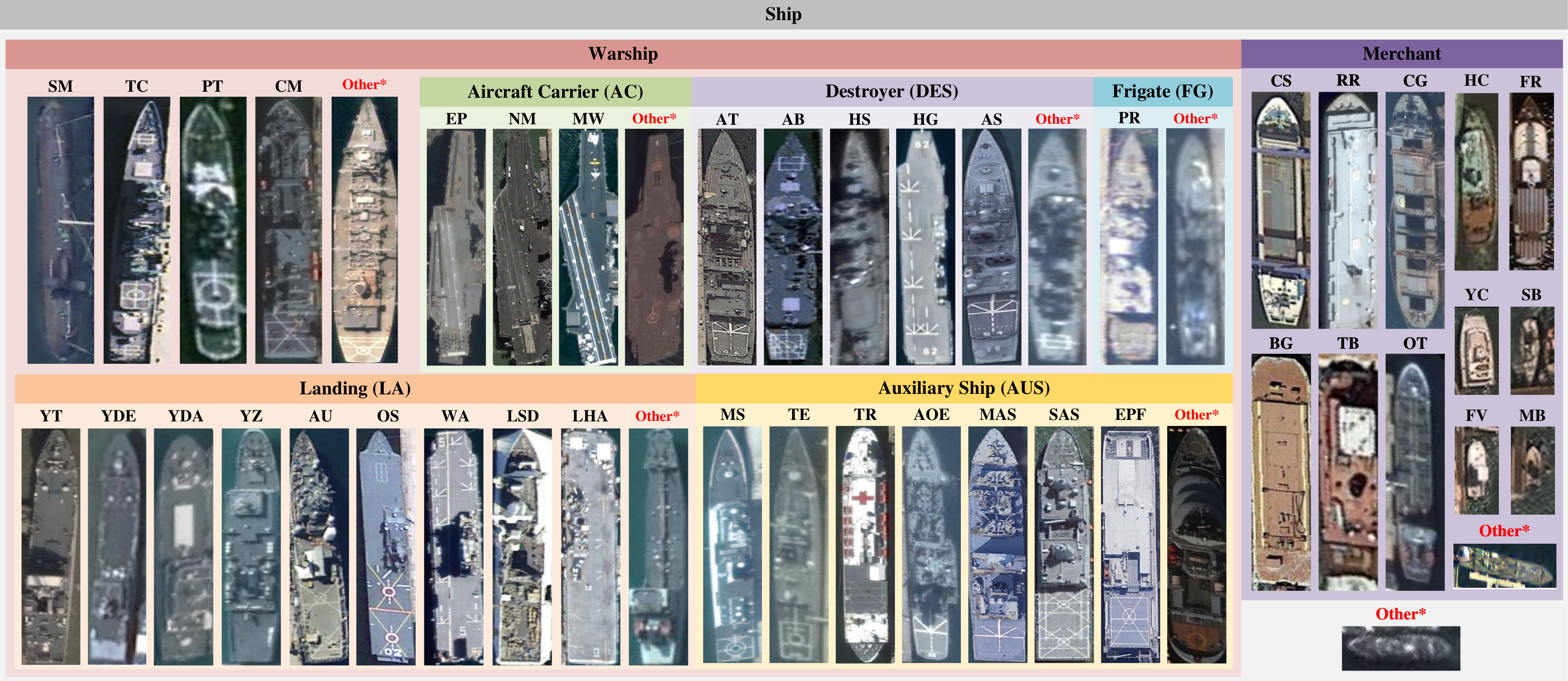}
    \caption{Hierarchical Label Structure of the ShipRSImageNet Dataset. The 41 fine-grained ship categories are organized into a four-level hierarchy by introducing 8 coarse-grained categories as parent nodes. Additionally, eight \textit{Other*} categories are included: Other Ship, Other Warship, Other Merchant, Other Aircraft Carrier, Other Destroyer, Other Frigate, Other Landing, and Other Auxiliary Ship.}
    \label{fig:ShipRSImageNet}
\end{figure*}

\noindent\textbf{ShipRSImageNet}. It contains 41 fine-grained ship categories, including Submarine (SM), Enterprise (EP), Nimitz (NM), Midway (MW), Atago (AT), Arleigh Burke (AB), Hatsuyuki (HS), Hyuga (HG), Asagiri (AS), Ticonderoga (TC), Perry (PR), Patrol (PT), YuTing (YT), YuDeng (YDE), YuDao (YDA), YuZhao (YZ), Austin (AU), Osumi(OS), Wasp (WA), LSD 41 (LSD), LHA, Commander(CM), Medical Ship (MS), Test Ship (TE), Training Ship (TR), AOE, Masyuu AS (MAS), Sanantonio AS (SAS), EPF, Container Ship (CS), RoRo (RR), Cargo (CG), Barge (BG), Tugboat (TB), Ferry (FR), Yacht (YC), Sailboat (SB), Fishing Vessel (FV), Oil Tanker (OT), Hovercraft (HC), and Motorboat (MB). These fine-grained categories are organized into a 4-level class hierarchy by adding 8 coarse-grained ancestor categories, as illustrated in Fig. \ref{fig:ShipRSImageNet}.

Specifically, EP, NM, and MW are categorized as subcategories of Aircraft Carrier (AC); AT, AB, HS, HG, and AS are classified under Destroyer (DES); PR is placed under Frigate (FG); YT, YDE, YDA, YZ, AU, OS, WA, LSD, and LHA are grouped as subcategories of Landing Ship (LA); and MS, TE, TR, AOE, MAS, SAS, and EPF are organized as subcategories of Auxiliary Ship (AUS). Then, SM, TC, PT, and CM, along with the previously introduced coarse-grained categories AC, DES, FG, LA, and AUS, are grouped into Warship. Meanwhile, CS, RR, CG, HC, FR, BG, TB, OT, YC, SB, FV, and MB are grouped into Merchant Ships. Finally, the most general category, Ship, is set as the root node, encompassing both Warship and Merchant Ships. Additionally, each coarse-grained category has an associated ``Other'' category, which covers instances that cannot be further identified due to low image resolution or those that do not belong to any predefined subcategory. In contrast to existing methods that treat these ``Other'' categories as mutually exclusive fine-grained categories, we reassign these instances to their corresponding coarse-grained categories.

\begin{figure}
    \centering
    \includegraphics[width=1.0\linewidth]{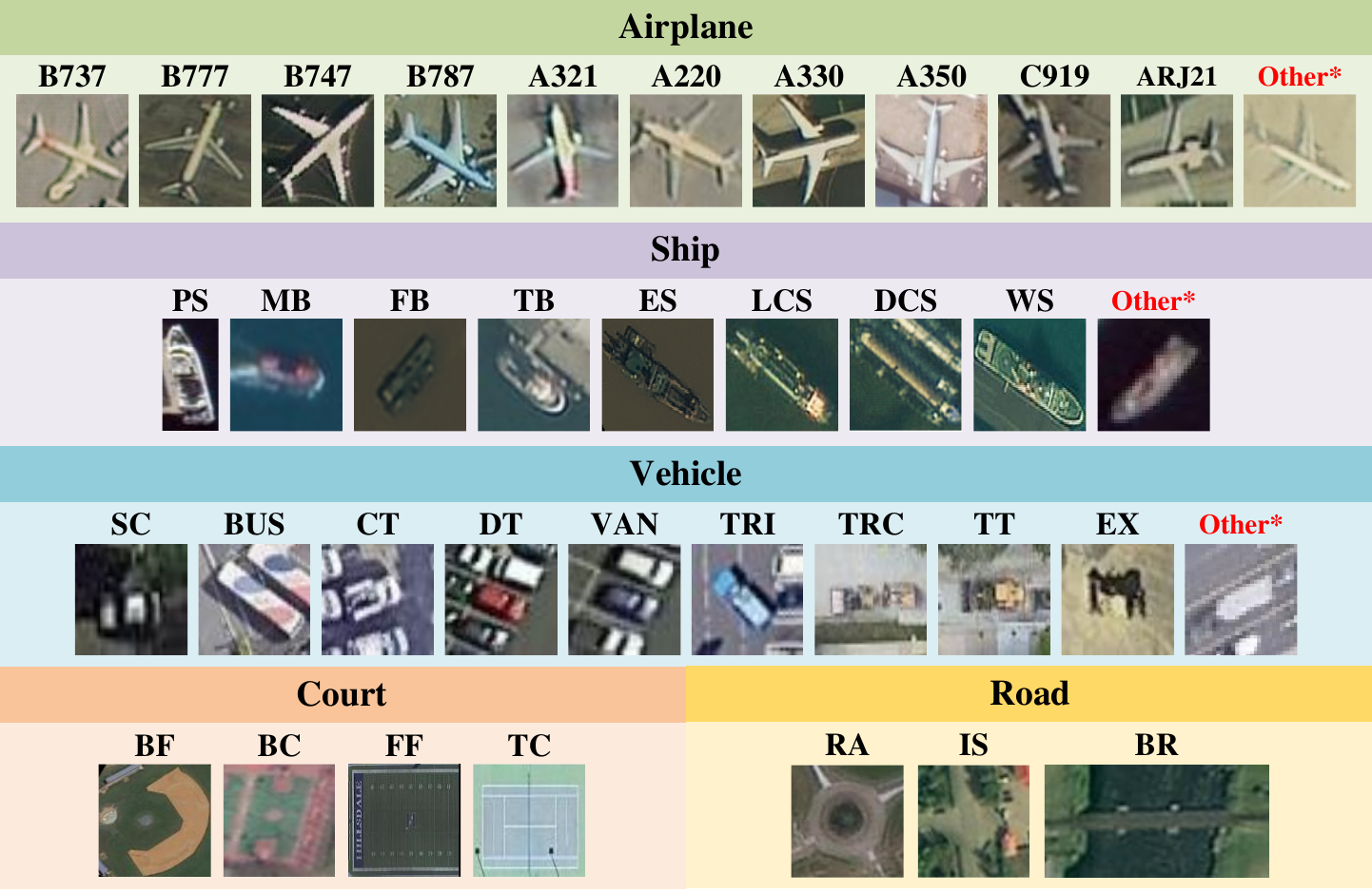}
    \caption{Hierarchical Label Structure of the FAIR1M Dataset. The 34 fine-grained categories are organized into a two-level hierarchy by introducing 5 coarse-grained categories as parent nodes. Additionally, three \textit{Other*} categories are included: Other Airplane, Other Ship, and Other Vehicle.}
    \label{fig:FARI1M}
    \vspace{-0.5cm}
\end{figure}

\noindent\textbf{FAIR1M-v1.0/v2.0}. Both FAIR1M-v1.0 and FAIR1M-v2.0 contain 34 fine-grained categories, including Boeing 737 (B737), Boeing 777 (B777), Boeing 747 (B747), Boeing 787 (B787), Airbus A321 (A321), Airbus A220 (A220), Airbus A330 (A330), Airbus A350 (A350), COMAC C919(C919), COMAC ARJ21 (ARJ21), passenger ship (PS), motorboat (MB), fishing boat (FB), tugboat (TB), engineering ship (ES), liquid cargo ship (LCS), dry cargo ship (DCS), warship (WS), small car (SC), bus (BUS), cargo truck (CT), dump truck (DT), van (VAN), trailer (TRI), tractor (TRC), truck tractor (TT), excavator (EX), baseball field (BF), basketball court (BC), football field (FF), tennis court (TC), roundabout (RA), intersection (IS), and bridge (BR). These fine-grained categories are organized into a 2-level class hierarchy, with 5 coarse-grained categories serving as their ancestors, as illustrated in Fig. \ref{fig:FARI1M}.

Specifically, B737, B777, B747, B787, A321, A220, A330, A350, C919, and ARJ21 are categorized as subcategories of Airplane; PS, MB, FB, TB, ES, LCS, DCS, and WS are classified under Ship; SC, BUS, CT, DT, VAN, TRI, TRC, TT, and EX are grouped as subcategories of Vehicle; BF, BC, FF, and TC are classified as subcategories of Court; and RA, IS, and BR are organized as subcategories of Road. Additionally, each of Airplane, Ship, and Vehicle includes an ``Other'' category, and we apply a reassignment procedure similar to that of ShipRSImageNet.

\section{Validation of Imbalanced Data Distribution}
\begin{figure}
    \vspace{-0.5cm}
    \centering
    \includegraphics[width=0.8\linewidth]{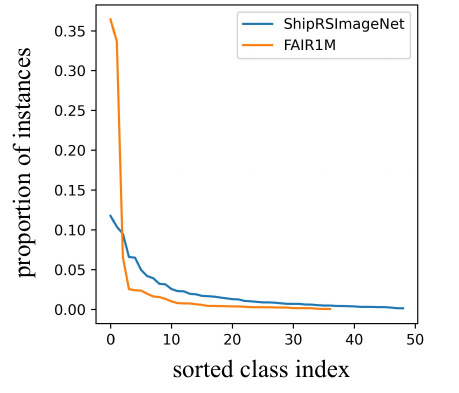}
    \caption{Instance Proportions per Category in the Training Sets of ShipRSImageNet and FAIR1M.}
    \label{fig:DataComparison}
    \vspace{-0.5cm}
\end{figure}

The experimental results in Section \ref{sec:exper_differentDatasets} reveal that, on the FAIR1M-v1.0 dataset, applying the hierarchical contrastive loss (HCL) alone leads to performance degradation, whereas the balanced hierarchical contrastive loss (BHCL) yields consistent improvements. We attribute this difference to the more severe long-tail distribution in FAIR1M-v1.0 compared to ShipRSImageNet, where HCL becomes dominated by instances from head classes. This hypothesis is supported by an analysis of instance proportions in the training sets of both datasets. As illustrated in Fig. \ref{fig:DataComparison}, the top three categories in FAIR1M-v1.0 (SC, VAN, and DT) account for over 70\% of all instances, significantly outnumbering the remaining categories and confirming the pronounced class imbalance.

\begin{figure}
    \vspace{-0.5cm}
    \centering
    \includegraphics[width=1.0\linewidth]{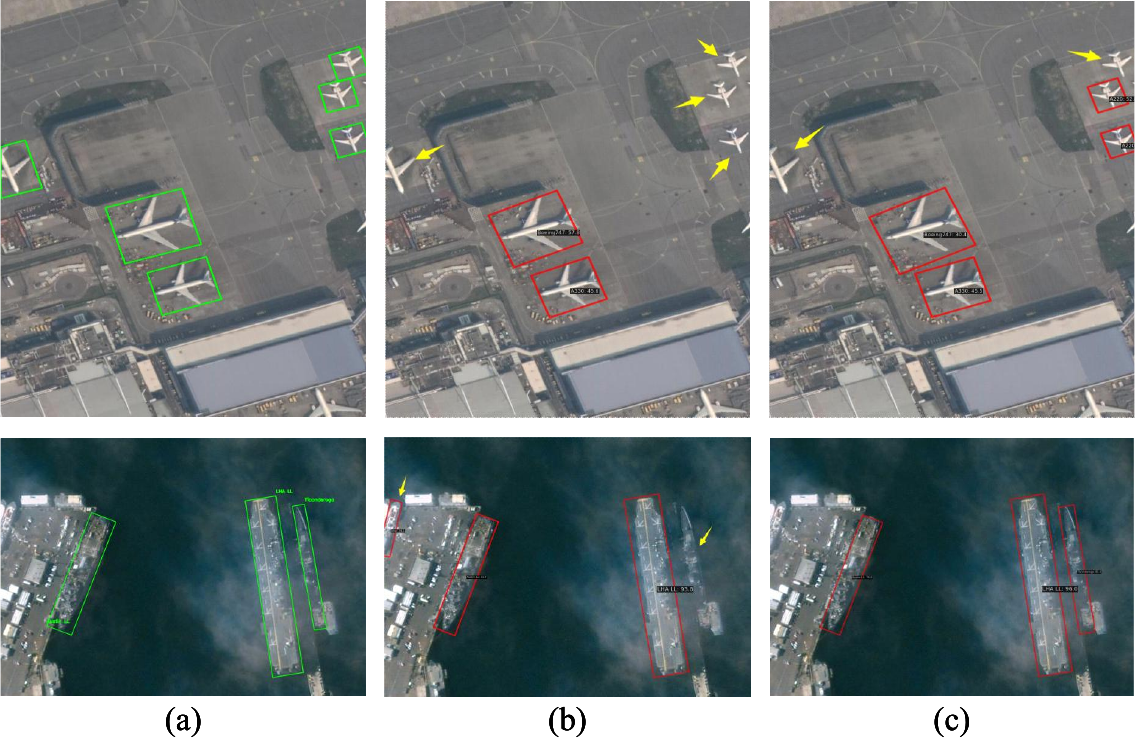}
    \caption{Comparison of Detection Results on the \mbox{ShipRSImageNet} (bottom) and \mbox{FAIR1M-v2.0} (top) Validation Sets: (a) Ground Truth, (b) \mbox{OrientedFormer}, (c) Proposed Method (Best Results).}
    \label{fig:detectionResults_More}
    \vspace{-0.5cm}
\end{figure}

\section{Fine-Grained Performance Comparisons with State-of-the-Art Methods}
In Section \ref{sec:sota_comparison}, we present overall performance comparisons between our proposed method and state-of-the-art approaches. Our method outperforms the second-best approach on the ShipRSImageNet, FAIR1M-v1.0, and FAIR1M-v2.0 datasets by $+1.1~AP_{50:95}$, $+0.35~AP_{50}$, and $+0.49~AP_{50}$, respectively. For a more detailed understanding of the performance differences, we provide fine-grained comparison results in Tables \ref{tab:SOTA_ShipRSImageNet_FineGrained}, \ref{tab:SOTA_FAIR1M1.0_FineGrained}, and \ref{tab:SOTA_FAIR1M2.0_FineGrained}. As evident from Tables \ref{tab:SOTA_FAIR1M1.0_FineGrained} and \ref{tab:SOTA_FAIR1M2.0_FineGrained}, our method achieves better performance over the compared methods, particularly in detecting fine-grained categories under Ship and Vehicle.

\section{Comparison of Additional Detection Results}
Section \ref{sec:vis_det_results} presents comparisons of detection results between OrientedFormer and our proposed method. In this section, we provide an additional set of comparative results. As illustrated in Fig. \ref{fig:detectionResults_More}, our method yields superior detection performance under misty conditions, indicating its greater robustness.

\begin{table*}[!b]\small
\caption{Comparison of Fine-Grained Detection Results with State-of-the-Art Methods on ShipRSImageNet. All methods employ a ResNet-50 backbone with input images resized to 1024 $\times$ 1024 pixels. Results for the compared methods, except RHINO and OrientedFormer, are adopted from \cite{SAFPN+APCL}.}\label{tab:SOTA_ShipRSImageNet_FineGrained}
\centering
\begin{tabular}{cccccccccc}
\toprule
Method & ReDet & ORCNN & PETDet & PCLDet & LSKNet & SAFPN+APCL & RHINO & OrientedFormer & \textbf{Ours}\\
\midrule
SM & 44.4 & \bf{69.0} & 46.3 & 66.9 & 62.6 & 68.8 & 45.5 & 43.7 & 43.5 \\
EP & 69.6 & 70.6 & 64.2 & \bf{77.5} & 72.4 & 75.7 & 69.0 & 73.5 & 69.0 \\
NM & 67.3 & 77.4 & 74.4 & 74.6 & 78.6 & 78.9 & 65.6 & \bf{82.8} & 73.3 \\
MW & 61.2 & \bf{84.8} & 74.2 & 53.3 & 60.4 & 71.2 & 77.6 & 74.1 & 81.8 \\
TC & 51.8 & \bf{83.6} & 67.9 & 82.9 & 82.8 & 82.0 & 62.9 & 67.6 & 70.1 \\
AT & 33.2 & 49.3 & 65.7 & 57.9 & 53.1 & 61.6 & 64.3 & 68.8 & \bf{70.6} \\
AB & 77.8 & 87.5 & 72.7 & 87.7 & 85.8 & \bf{87.8} & 69.3 & 77.5 & 75.9 \\
HS & 23.2 & 60.0 & 66.1 & 50.9 & 41.9 & 56.6 & 69.1 & \bf{70.6} & 68.9 \\
HG & 84.1 & \bf{100.0} & 78.8 & 97.3 & 97.4 & 98.2 & 80.5 & 81.7 & 84.9 \\
AS & 4.4 & 18.6 & 51.3 & 17.9 & 18.7 & 37.8 & \bf{66.0} & 61.0 & 60.9 \\
PR & 75.3 & 89.6 & 70.9 & 88.6 & 87.2 & \bf{90.2} & 68.2 & 73.8 & 73.4 \\
PT & 21.6 & 59.8 & 39.0 & 41.0 & \bf{72.0} & 56.2 & 48.8 & 54.2 & 50.7 \\
YT & 48.3 & 61.7 & 59.7 & 59.5 & \bf{70.3} & 66.2 & 58.6 & 50.9 & 65.2 \\
YDE & 3.6 & \bf{77.1} & 62.6 & 41.8 & 44.7 & 59.9 & 63.4 & 67.8 & 68.0 \\
YDA & 51.4 & 54.5 & 54.9 & 39.4 & \bf{92.4} & 64.5 & 63.9 & 57.6 & 62.0 \\
YZ & 31.2 & 67.6 & 69.4 & 73.6 & 73.5 & 58.2 & 76.4 & \bf{81.8} & 74.2 \\
AU & 59.2 & 50.8 & 72.8 & 68.8 & 67.3 & 64.0 & 65.4 & \bf{76.5} & 68.4 \\
OS & 81.8 & 85.7 & 85.6 & \bf{100.0} & 97.4 & \bf{100.0} & 87.1 & 91.8 & 92.7 \\
WA & 37.3 & 71.9 & 83.9 & 92.7 & \bf{97.4} & 89.4 & 90.9 & 90.9 & 90.8 \\
LSD & 67.7 & 67.2 & 66.9 & \bf{74.7} & 58.9 & 69.2 & 63.1 & 67.8 & 70.5 \\
LHA & 71.8 & \bf{89.1} & 64.6 & 88.2 & 87.3 & 88.4 & 69.0 & 67.8 & 71.7 \\
CM & 64.0 & 81.6 & 79.8 & \bf{84.6} & 77.1 & 80.4 & 72.6 & 81.5 & 81.8 \\
MS & 29.2 & 51.5 & 76.6 & 74.7 & 62.2 & 63.3 & 69.8 & 72.5 & \bf{77.6} \\
TE & 10.4 & 37.9 & 54.2 & 36.6 & 46.0 & 41.9 & 54.0 & \bf{71.5} & 69.8 \\
TR & 30.3 & 67.4 & 68.8 & \bf{92.9} & 73.2 & 72.3 & 79.9 & 79.1 & 79.5 \\
MAS & 13.6 & 60.9 & 81.2 & 74.3 & 66.9 & 65.1 & 85.9 & \bf{92.7} & 90.9 \\
AOE & 7.8 & 11.4 & 81.8 & 27.0 & 36.2 & 27.3 & 84.3 & \bf{89.6} & 89.5 \\
SAS & 47.3 & 68.7 & 75.8 & 72.4 & 64.7 & 70.4 & 75.2 & 72.0 & \bf{76.8} \\
EPF & \bf{82.7} & 81.4 & 65.2 & 76.7 & 67.7 & 82.1 & 72.8 & 78.5 & 79.6 \\
CS & 48.4 & 50.3 & 44.4 & 50.1 & 47.5 & \bf{56.1} & 46.7 & 47.5 & 47.4 \\
RR & 73.4 & 76.1 & 62.4 & 82.1 & 74.1 & \bf{83.1} & 69.0 & 64.9 & 68.8 \\
CG & 58.3 & \bf{63.6} & 50.5 & 58.5 & 55.5 & \bf{63.6} & 49.6 & 53.9 & 58.3 \\
BG & 14.6 & 5.7 & 17.3 & 9.9 & 11.0 & 13.7 & 16.6 & 24.6 & \bf{33.3} \\
TB & 46.1 & 51.0 & 38.2 & 55.6 & \bf{57.0} & 54.3 & 37.4 & 37.7 & 38.0 \\
FR & 23.0 & 22.2 & 34.1 & 29.1 & 19.8 & 22.7 & 32.7 & 38.3 & \bf{41.3} \\
YC & 64.6 & 69.9 & 56.6 & 73.5 & \bf{73.7} & 71.0 & 53.1 & 59.7 & 56.4 \\
SB & 15.5 & 15.8 & 9.0 & 16.6 & 17.1 & \bf{18.9} & 8.7 & 16.2 & 17.5 \\
FV & 29.3 & \bf{34.3} & 18.4 & 25.0 & 29.7 & 31.2 & 24.2 & 24.4 & 28.5 \\
OT & 41.0 & 43.8 & 46.9 & \bf{65.5} & 52.0 & 45.8 & 40.8 & 44.3 & 50.0 \\
HC & 64.2 & 55.9 & 47.5 & \bf{65.8} & 50.5 & 60.4 & 44.6 & 47.8 & 53.5 \\
MB & 18.8 & 16.1 & 7.1 & \bf{21.0} & 17.6 & 20.3 & 8.6 & 11.0 & 11.6 \\
\midrule
$AP_{50:95}$ & 45.1 & 59.5 & 58.7 & 61.6 & 61.0 & 62.7 & 59.8 & 63.2 & \bf{64.3} \\
\bottomrule
\end{tabular}
\end{table*}

\begin{table*}\small
\caption{Comparison of Fine-Grained Detection Results with State-of-the-Art Methods on FAIR1M-v1.0. All methods employ a ResNet-50 backbone with input images resized to 1024 $\times$ 1024 pixels. Results for the compared methods, except OrientedFormer, are adopted from \cite{DRNet}.}\label{tab:SOTA_FAIR1M1.0_FineGrained}
\centering
\begin{tabular}{cccccccccc}
\toprule
Method & FRCNN & RoITrans & ORCNN & PCLDet & SFRNet & PETDet & DRNet & OrientedFormer & \textbf{Ours} \\
\midrule
B737 & 33.94 & 39.15 & 35.17 & 35.96 & 39.70 & \bf{41.05} & 39.05 & 38.46 & 38.64 \\
B747 & 84.25 & 84.72 & 85.17 & \bf{85.50} & 84.44 & 82.23 & 84.91 & 83.86 & 83.57 \\
B777 & 16.38 & 14.82 & 14.57 & 15.15 & 17.79 & 22.04 & 16.42 & \bf{24.22} & 21.65 \\
B787 & 47.61 & 48.88 & 47.68 & 47.97 & 48.96 & \bf{51.21} & 49.13 & 41.30 & 43.02 \\
C919 & 14.44 & 19.49 & 11.68 & 16.97 & 21.18 & 25.66 & \bf{52.50} & 17.12 & 23.36 \\
A220 & 47.40 & 50.31 & 39.05 & 45.64 & 48.38 & 51.83 & \bf{68.26} & 48.09 & 47.06 \\
A321 & 68.82 & 70.16 & 39.05 & 69.14 & \bf{71.25} & 69.53 & 66.98 & 66.62 & 66.07 \\
A330 & \bf{72.71} & 70.34 & 68.60 & 71.54 & 72.06 & 70.86 & 39.05 & 65.49 & 67.39 \\
A350 & \bf{76.53} & 72.19 & 70.21 & 71.50 & 74.16 & 73.70 & 70.27 & 74.08 & 70.09 \\
ARJ21 & 26.59 & 33.72 & 25.32 & 38.68 & 31.47 & 36.90 & \bf{45.45} & 35.32 & 34.74 \\
PS & 11.03 & 12.62 & 13.77 & 17.43 & \bf{17.44} & 13.29 & 16.74 & 15.86 & 14.19 \\
MB & 51.22 & 55.98 & 60.42 & 56.85 & 60.61 & 61.94 & 58.30 & 63.95 & \bf{66.40} \\
FB & 6.41 & 6.12 & 9.10 & 7.88 & 8.50 & 8.90 & 8.08 & 9.74 & \bf{10.11} \\
TB & 34.19 & 35.31 & 36.83 & 36.70 & 34.87 & 37.88 & 32.52 & 36.95 & \bf{39.04} \\
ES & 9.41 & 9.27 & 11.32 & 10.97 & \bf{12.55} & 10.93 & 11.97 & 11.40 & 12.21 \\
LCS & 15.17 & 15.95 & 21.86 & 19.91 & 19.71 & 22.05 & 22.07 & 22.10 & \bf{22.87} \\
DCS & 32.26 & 34.15 & 38.22 & \bf{40.05} & 38.25 & 36.82 & 37.74 & 39.01 & 39.41 \\
WS & 11.27 & 15.29 & 22.67 & 23.80 & 22.00 & 23.98 & 24.79 & \bf{28.62} & 28.40 \\
SC & 54.56 & 57.55 & 57.62 & 56.78 & 57.73 & 68.81 & 58.39 & \bf{70.62} & \bf{70.62} \\
BUS & 22.94 & 26.43 & 24.40 & 35.10 & 32.37 & 18.33 & 35.51 & 36.79 & \bf{38.56} \\
CT & 37.74 & 39.38 & 40.84 & 42.19 & 41.01 & 42.17 & 42.90 & 45.00 & \bf{46.26} \\
DT & 41.69 & 44.95 & 45.20 & 45.83 & 46.69 & 47.15 & 47.23 & 49.47 & \bf{50.40} \\
VAN & 48.23 & 53.69 & 54.01 & 52.01 & 54.08 & 65.48 & 53.59 & 70.86 & \bf{70.92} \\
TRI & 12.46 & 10.95 & 15.46 & \bf{16.34} & 15.75 & 11.93 & 15.80 & 11.54 & 13.16 \\
TRC & 2.44 & 2.13 & 2.37 & 6.43 & \bf{7.09} & 2.09 & 4.26 & 1.99 & 4.85 \\
EX & 11.35 & 10.99 & 13.55 & \bf{18.08} & 15.97 & 8.51 & 17.42 & 16.28 & 17.28 \\
TT & 0.32 & 0.60 & 0.24 & 0.61 & 0.37 & 0.41 & 1.05 & 0.60 & \bf{1.29} \\
BC & 45.18 & 46.93 & 48.18 & 48.45 & 49.43 & 42.64 & 50.43 & \bf{50.86} & 47.56 \\
TC & 77.75 & 79.29 & 78.45 & 78.13 & 79.17 & 78.60 & \bf{82.25} & 79.81 & 79.04 \\
FF & 52.05 & 56.72 & 60.79 & 62.55 & 59.90 & \bf{63.51} & 60.53 & 54.97 & 55.97 \\
BF & 87.19 & 87.21 & \bf{88.43} & 88.23 & 87.90 & 87.40 & 86.96 & 85.59 & 86.61 \\
IS & 58.71 & 58.21 & 57.90 & 59.01 & 59.48 & 57.48 & 58.51 & \bf{60.42} & 59.52 \\
RA & 19.38 & 21.98 & 17.57 & 20.21 & 22.05 & 20.82 & \bf{24.29} & 17.97 & 16.58 \\
BR & 20.76 & 23.31 & 28.63 & 31.67 & \bf{32.76} & 22.57 & 26.58 & 29.61 & 29.68 \\
\midrule
$AP_{50}$ & 36.83 & 38.49 & 38.85 & 40.39 & 40.74 & 40.55 & 40.87 & 41.31 & \bf{41.66} \\
\bottomrule
\end{tabular}
\end{table*}

\begin{table*}\small
\caption{Comparison of Fine-Grained Detection Results with State-of-the-Art Methods on FAIR1M-v2.0. All methods employ a ResNet-50 backbone with input images resized to 1024 $\times$ 1024 pixels. Results for the compared methods are adopted from \cite{DRNet}.}\label{tab:SOTA_FAIR1M2.0_FineGrained}
\centering
\begin{tabular}{cccccccccc}
\toprule
Method & FRCNN & RoITrans & GVertex & ORCNN & PCLDet & SFRNet & PETDet & DRNet & \textbf{Ours} \\
\midrule
B737 & 42.84 & 47.76 & 42.95 & 44.24 & 42.77 & 45.29 & \bf{48.96} & 43.28 & 45.04 \\
B747 & 93.28 & 94.13 & 92.98 & 93.27 & 93.44 & 92.97 & \bf{94.57} & 93.68 & 93.26 \\
B777 & 35.63 & 38.62 & 37.59 & 36.81 & 36.96 & 40.87 & \bf{43.24} & 39.87 & 38.72 \\
B787 & 62.19 & 62.21 & 57.19 & 59.27 & 61.60 & 61.21 & \bf{64.50} & 63.72 & 62.13 \\
C919 & 3.58 & 12.18 & 6.31 & 6.35 & 12.79 & 15.96 & \bf{26.89} & 16.49 & 11.91 \\
A220 & 53.51 & 54.96 & 53.02 & 55.29 & 52.53 & 54.17 & \bf{57.45} & 54.20 & 57.14 \\
A321 & 68.28 & 69.07 & 67.11 & 70.25 & 69.76 & 69.36 & \bf{74.41} & 68.07 & 70.61 \\
A330 & 58.65 & 60.11 & 57.98 & 61.05 & \bf{64.11} & 61.50 & 62.84 & 59.82 & 62.56 \\
A350 & 63.46 & 66.17 & 65.02 & 68.04 & 68.18 & 68.26 & 68.45 & \bf{70.25} & 68.76 \\
ARJ21 & 10.80 & 15.34 & 12.35 & 10.40 & 13.52 & 15.68 & 11.99 & \bf{18.92} & 17.62 \\
PS & 10.80 & 13.51 & 12.75 & 14.12 & 14.93 & 14.45 & 14.13 & \bf{17.00} & 14.15 \\
MB & 51.82 & 57.76 & 56.51 & 61.61 & 58.04 & 61.47 & 62.88 & 62.03 & \bf{65.56} \\
FB & 12.65 & 20.89 & 15.99 & 26.14 & 25.12 & 26.58 & 24.61 & \bf{29.38} & 26.22 \\
TB & 28.35 & 30.78 & 27.63 & 30.55 & 30.34 & 30.56 & 31.59 & \bf{33.01} & 32.45 \\
ES & 11.45 & 15.12 & 13.23 & 15.97 & 17.03 & 17.89 & 16.53 & \bf{18.39} & 17.75 \\
LCS & 32.95 & 45.49 & 40.72 & 49.45 & 50.76 & 48.74 & 49.88 & 50.85 & \bf{53.01} \\
DCS & 35.34 & 49.50 & 42.42 & 51.95 & 52.11 & 50.79 & 49.46 & 52.11 & \bf{54.26} \\
WS & 13.88 & 29.06 & 20.95 & 33.01 & 30.22 & 31.98 & 28.92 & 36.78 & \bf{38.14} \\
SC & 52.23 & 58.68 & 46.62 & 56.41 & 56.94 & 56.77 & 70.17 & 58.57 & \bf{71.91} \\
BUS & 24.12 & 30.25 & 26.47 & 31.81 & 34.10 & 32.58 & 31.31 & \bf{44.71} & 35.58 \\
CT & 45.51 & 50.35 & 44.21 & 51.29 & 49.85 & 50.93 & 52.91 & 53.10 & \bf{55.26} \\
DT & 44.54 & 48.80 & 41.74 & 48.74 & 49.71 & 49.00 & 51.14 & 50.36 & \bf{54.35} \\
VAN & 47.00 & 52.62 & 40.20 & 52.87 & 51.24 & 52.77 & 68.30 & 52.87 & \bf{72.12} \\
TRI & 8.63 & 13.18 & 12.95 & 15.89 & 15.81 & \bf{17.56} & 12.94 & 16.54 & 16.45 \\
TRC & 1.12 & 2.73 & 0.85 & 1.55 & 2.46 & 2.47 & 0.47 & 1.83 & \bf{7.48} \\
EX & 8.43 & 13.79 & 12.12 & 15.96 & 17.52 & 16.20 & 13.34 & \bf{20.85} & 17.70 \\
TT & 5.93 & 18.57 & 6.24 & 8.06 & 7.60 & 29.35 & 9.96 & 27.34 & \bf{36.33} \\
BC & 55.31 & 54.71 & 55.08 & 59.77 & 60.65 & 60.97 & 53.89 & \bf{62.89} & 56.92 \\
TC & 85.76 & 86.40 & 86.70 & 88.11 & 87.45 & 89.35 & 87.76 & \bf{90.47} & 88.05 \\
FF & 60.48 & 63.02 & 58.29 & 63.70 & 64.83 & 61.49 & \bf{66.38} & 66.30 & 58.82 \\
BF & 91.11 & 90.93 & 89.96 & 91.63 & \bf{92.02} & 90.56 & 91.29 & 91.01 & 90.45 \\
IS & 63.04 & 65.17 & 64.34 & 64.37 & \bf{65.85} & 65.11 & 64.67 & 65.62 & 64.69 \\
RA & 29.98 & 33.78 & 32.30 & 31.52 & 30.72 & \bf{35.13} & 33.06 & 34.13 & 28.00 \\
BR & 25.90 & 34.07 & 30.62 & 36.47 & \bf{37.44} & 35.05 & 28.57 & 34.87 & 32.64 \\
\midrule
$AP_{50}$ & 39.37 & 44.11 & 40.33 & 44.29 & 44.66 & 45.68 & 46.10 & 47.04 & \bf{47.53} \\
\bottomrule
\end{tabular}
\end{table*}


\end{document}